\documentclass{article}

\usepackage{microtype}
\usepackage{graphicx}
\usepackage{subcaption}
\usepackage{booktabs} 

\usepackage{hyperref}

\usepackage[preprint]{icml2026}

\usepackage{amsmath}
\usepackage{amssymb}
\usepackage{mathtools}
\usepackage{amsthm}

\usepackage[capitalize,noabbrev]{cleveref}

\theoremstyle{plain}

\theoremstyle{definition}

\theoremstyle{remark}

\usepackage[textsize=tiny]{todonotes}

\usepackage{fontawesome}
\usepackage{graphicx}
\usepackage{xspace}
\usepackage{amsmath}
\usepackage[dvipsnames]{xcolor}
\usepackage{multirow}
\usepackage{booktabs}
\usepackage{lipsum}
\usepackage{subcaption}
\usepackage{hyperref}
\usepackage{url}
\usepackage{amssymb}
\usepackage{tcolorbox}
\usepackage{titlesec}
\usepackage{titletoc}
\usepackage{lineno}
\usepackage[table]{xcolor}

\newcommand{\method}{\textsc{dVoting}\xspace}

\definecolor{map_red}{RGB}{239,99,75}
\definecolor{map_blue}{RGB}{99,113,250}
\definecolor{map_green}{RGB}{0,180,139}
\definecolor{map_yellow}{RGB}{229,157,35}
\definecolor{map_gray}{RGB}{165,165,165}
\definecolor{link}{RGB}{229,158,221}
\definecolor{mygreen}{RGB}{93,173,85}
\definecolor{myred}{RGB}{192,57,43}

\hypersetup{
  colorlinks=true,
  linkcolor={red!70!black},
  citecolor=link,
  urlcolor=cyan,
}

\icmltitlerunning{\method: Fast Voting for dLLMs}

\begin{document}

\twocolumn[
  \icmltitle{\method: Fast Voting for dLLMs}



  \icmlsetsymbol{corr}{*}

  \begin{icmlauthorlist}
    \icmlauthor{Sicheng Feng}{nus}
    \icmlauthor{Zigeng Chen}{nus}
    \icmlauthor{Xinyin Ma}{nus}
    \icmlauthor{Gongfan Fang}{nus}
    \icmlauthor{Xinchao Wang}{nus,corr}

  \end{icmlauthorlist}

  \icmlaffiliation{nus}{Department of Electrical and Computer Engineering, National University of Singapore, Singapore}

  \icmlcorrespondingauthor{Xinchao Wang}{xinchao@nus.edu.sg}

  \icmlkeywords{Machine Learning, ICML}

  \vskip 0.3in
]



\printAffiliationsAndNotice{}  

\begin{abstract}
Diffusion Large Language Models (dLLMs) represent a new paradigm beyond autoregressive modeling, offering competitive performance while naturally enabling a flexible decoding process.
%
Specifically, dLLMs can generate tokens at arbitrary positions in parallel, endowing them with significant potential for parallel test-time scaling, which was previously constrained by severe inefficiency in autoregressive modeling.
In this work, we introduce \method, a fast voting technique that boosts reasoning capability without training, with only an acceptable extra computational overhead.
%
\method is motivated by the observation that, across multiple samples for the same prompt, token predictions remain largely consistent, whereas performance is determined by a small subset of tokens exhibiting cross-sample variability. Leveraging the arbitrary-position generation capability of dLLMs, \method performs iterative refinement by sampling, identifying uncertain tokens via consistency analysis, regenerating them through voting, and repeating this process until convergence.
%
Extensive evaluations demonstrate that \method consistently improves performance across various benchmarks.
It achieves gains of $6.22\%$-$7.66\%$ on GSM8K, $4.40\%$-$7.20\%$ on MATH500, $3.16\%$-$14.84\%$ on ARC-C, and $4.83\%$-$5.74\%$ on MMLU. 
%
Our code is available at \url{https://github.com/fscdc/dVoting}
\end{abstract}

\section{Introduction}
\label{sec:intro}

\begin{figure*}[t]
\centering
  \includegraphics[width=\linewidth]{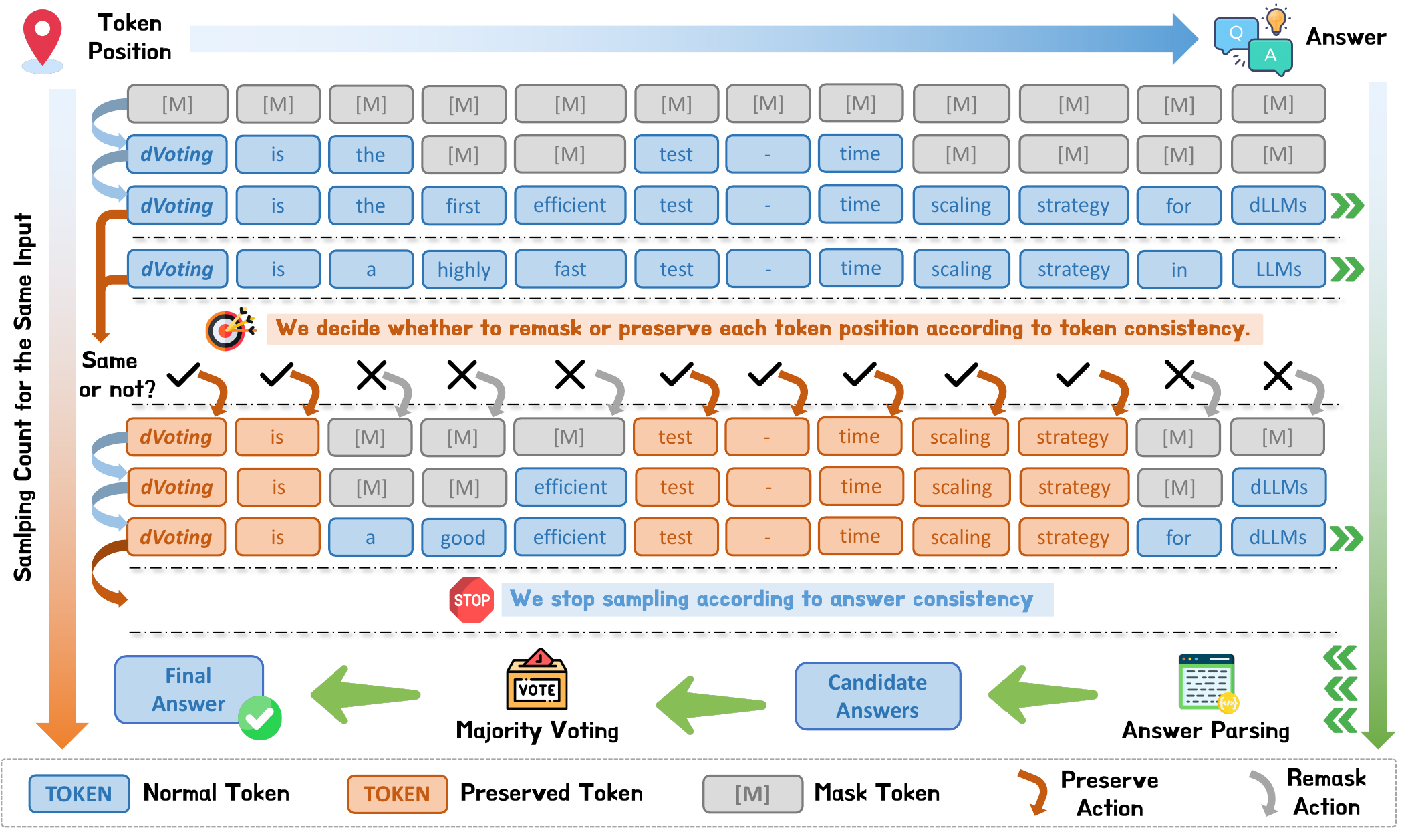} \\ 
\vspace{-2mm}
\caption{Overview of \method. For each prompt, our \method preserves consistent tokens in previous generations and remasks the remaining tokens to initiate subsequent sampling, and terminates the process early when candidate answers satisfy consistent criteria.}
\vspace{-4mm}
\label{fig:overview}
\end{figure*}

Diffusion large language models (dLLMs)~\citep{yi2024diffusion,zhang2025survey,nie2025large,ye2025dream,yu2025dimple,bie2025llada2} have recently emerged as a competitive alternative to autoregressive LLMs~\citep{achiam2023gpt,bai2023qwen,dubey2024llama}, demonstrating strong performance and extending beyond open-source settings to closed-source models such as Gemini-Diffusion, Seed-Diffusion~\citep{song2025seed}, and Mercury~\citep{khanna2025mercury}. Through an iterative unmask-and-remask process, dLLMs enable parallel decoding in a flexible order beyond left-to-right token generation in autoregressive LLMs, offering substantial flexibility and potential at test time.

In dLLMs, recent efforts~\citep{zhao2025d1,tang2025wd1,zhao2025inpainting,wang2025revolutionizing,yang2025taming,huang2025reinforcing} for reasoning enhancement mainly focus on training time, primarily through reinforcement learning (RL)~\citep{ouyang2022training,shao2024deepseekmath,feng2025rewardmap}.
These RL approaches introduce novel training schemes and achieve notable performance improvements. 
Complementary to these training-time approaches, another line of work explores enhancing reasoning by operating directly on the decoding process at test time.

Inspired by the success of test-time scaling in LLMs~\citep{muennighoff2025s1,snell2024scaling,wang2025sampling,sun2024fast,xu2025phi} and recent findings that RL primarily improves sampling efficiency rather than intrinsic capability~\citep{yue2025does,chen2025pass}, we instead focus on inference.
To date, a few works have explored test-time scaling for reasoning enhancement in dLLMs. HEX~\citep{lee2025test} activates implicit semi-autoregressive experts by varying block sizes, while RFG~\citep{chen2025rfg} guides generation at the logit level using an additional fine-tuned model.
Although these methods can boost reasoning performance, they typically involve increased inference-time computation. Our goal is to reduce the substantial redundancy inherent in the voting paradigm.

\begin{figure*}[t]
\centering
\includegraphics[width=\linewidth]{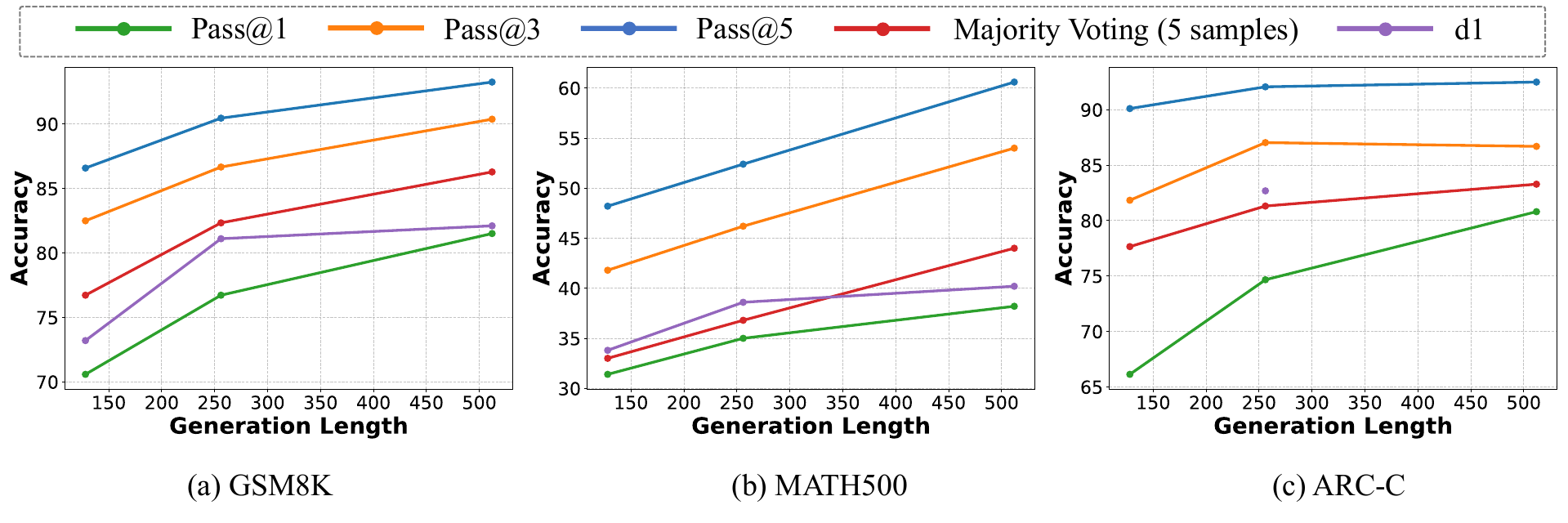}  \\
\vspace{-3mm}
\caption{Empirical study on LLaDA-8B-Instruct. We report the performance of five strategies: (1) \textcolor[HTML]{2ca02c}{Pass@1}; (2) \textcolor[HTML]{ff7f0e}{Pass@3}; (3) \textcolor[HTML]{1f77b4}{Pass@5}; (4) \textcolor[HTML]{d62728}{Majority voting ($5$ samples)}, which denotes the results of standard test-time scaling strategies; and (5) \textcolor[HTML]{9467bd}{d1}, which represents the performance of RL-enhanced models. We report the results of GSM8K, MATH500, and ARC-C in (a), (b), and (c), respectively.}
\label{fig:empirical-study}
\vspace{-4mm}
\end{figure*}

Based on the consistency analysis, we further identify a simple yet crucial observation that repeated tokens frequently appear across multiple samples for the same input, as illustrated in Figure~\ref{fig:empirical-study}(b) and quantified in Table~\ref{tab:observation}. Furthermore, we connect this observation with the remasking mechanism of dLLMs which allows the model to mask and regenerate an arbitrary number of decoded tokens at arbitrary positions within a sequence\footnote{\textit{e.g.}, ``dLLMs are very efficient at test time.'' $\rightarrow$ ``dLLMs [MASK] [MASK] [MASK] at [MASK] time.'' $\rightarrow$ ``dLLMs can be fast at inference time.''}, making it well-suited to reduce the redundancy revealed by this observation. Accordingly, we propose a simple yet fast voting strategy (Figure~\ref{fig:overview}). Specifically, we identify uncertain tokens based on token consistency, iteratively remask and regenerate them for refinement, and finally aggregate candidate answers via voting.

We conduct extensive experiments across various reasoning benchmarks on two popular dLLMs (LLaDA~\citep{nie2025large} \& Dream~\citep{ye2025dream}) to evaluate the effectiveness of our method. The results demonstrate that our method boosts the performance across all benchmarks covering mathematical, scientific, and general reasoning. For instance, our method yields performance gains of $6.22\%\text{–}7.66\%$ on GSM8K~\citep{cobbe2021gsm8k} and $4.40\%\text{–}7.20\%$ on MATH500~\citep{lightman2023let} on LLaDA. Our method further achieves a leading performance–efficiency trade-off. Additionally, we present the robustness of our method under various configurations.

%
In conclusion, we propose \method, a training-free, simple yet effective voting strategy that boosts performance. Based on the empirical study, we identify and quantify a key observation that many tokens are repeatedly generated across multiple sampling runs, and relate this to the remasking mechanism of dLLMs. Accordingly, we introduce the remask sampling strategy, which iteratively remasks and regenerates selected tokens to obtain multiple candidate generations and aggregates via voting to enhance performance. Extensive evaluations demonstrate the effectiveness of \method. Together, \method establishes the first baseline and provides a foundation for efficient test-time scaling in dLLMs, further unlocking their potential at test time.

\section{Related Work}
\label{sec:related_work}

\noindent \textbf{Overview of Diffusion Language Models.}
Diffusion models~\citep{ho2020denoising,song2019generative,song2020denoising} have shown strong generative capabilities in continuous domains (\textit{e.g.}, image~\citep{rombach2022high,peebles2023scalable}, video~\citep{ho2022video,brooks2024video}, and audio~\citep{liu2023audioldm,evans2024fast}), while their extension to language modeling remains challenging due to the discrete nature of text. Recent works~\citep{austin2021structured,sahoo2024simple,lou2023discrete,zheng2024masked,cheng2025sdar} address this issue by formulating diffusion processes directly over token spaces, typically through masked token prediction, which enables parallel generation and relaxes the strict autoregressive constraint. Building on this paradigm, dLLMs~\citep{nie2025large,ye2025dream,khanna2025mercury,song2025seed,bie2025llada2} have demonstrated competitive performance compared to autoregressive models at the billion-parameter scale, indicating the practical viability of diffusion for language generation. Moreover, diffusion language models have been increasingly explored in advanced settings such as reasoning~\citep{zhu2025llada,zhao2025d1,tang2025wd1,lin2025boundary}, multimodal generation~\cite{yang2025mmada,li2025lavida,yu2025dimple,you2025llada}, and code synthesis~\citep{gong2025diffucoder,khanna2025mercury,pengcontributors}, reflecting the growing scope and maturity of this research direction~\citep{yu2025discrete,li2025survey}. 

\noindent \textbf{Test-Time Scaling in Language Models.}
Test-time scaling~\citep{welleck2024decoding,snell2024scaling,muennighoff2025s1,openaio1} has emerged as an effective alternative to training-time scaling, aiming to elicit stronger reasoning capabilities by allocating additional computation during inference. In autoregressive language models, test-time scaling has been extensively studied through techniques such as chain-of-thought (CoT) prompting~\citep{wei2022chain,kojima2022large,zhou2022least}, best-of-N sampling~\citep{sun2024fast,wang2025sampling,xu2025phi}, and self-consistency~\citep{wang2022self,wang2024make,aggarwal2023let}. However, test-time scaling in dLLMs remains relatively underexplored. Recent studies provide initial evidence of its potential: HEX~\citep{lee2025test} shows that aggregating diverse masking schedules during inference can substantially improve performance, while RFG~\citep{chen2025rfg} introduces guidance from an additional fine-tuned model by operating at the logits level. These methods provide valuable references for test-time scaling in dLLMs. However, test-time scaling efficiency~\citep{sun2024fast,xu2025phi,feng2025efficient,ma2024non,zhu2024path} is equally critical, as it directly addresses the substantial inference cost and determines the practicality of such approaches in real-world settings.
To date, efficient test-time scaling for dLLMs remains largely unexplored; this gap is the primary focus of our work.

\section{Preliminaries}
\label{sec:preliminaries}

We anchor our preliminaries in continuous-time diffusion language models defined over discrete vocabularies, with masked diffusion language models (MDLMs) serving as the main instantiation. MDLMs generate text by iteratively reconstructing a partially corrupted sequence rather than producing tokens in an autoregressive order. Starting from an initial clean sequence $x_{0}$, the model constructs a noised version $x_{t}$ by independently masking each position with intensity $t \in [0,1]$. Formally, the corruption distribution is
\begin{equation*}
q(x_t \mid x_0)
=
\prod_{i=1}^{L}
\left[
\begin{aligned}
&(1-t)\,\delta(x_t^{i} = x_0^{i}) \\
&\quad + t\,\delta(x_t^{i} = \text{[MASK]})
\end{aligned}
\right].
\end{equation*}
Once a corrupted sequence is obtained, a denoising network $p_\theta$ attempts to infer the original tokens at the masked locations. Since all masked positions are conditionally independent given $x_t$, the recovery model factorizes as
\begin{equation*}
p_\theta(x_0 \mid x_t)
=
\prod_{i : x_t^{i} = \text{[MASK]}}
p_\theta(x_0^{i} \mid x_t).
\end{equation*}
Training focuses on masked tokens, with the model learning to reconstruct original content from corrupted sequences. The resulting reconstruction loss serves as a surrogate objective that upper-bounds the negative log-likelihood.

\begin{figure*}[t]
\centering
\resizebox{0.995\linewidth}{!}{
\begin{tabular}{cc} 
  \includegraphics[width=0.48\linewidth]{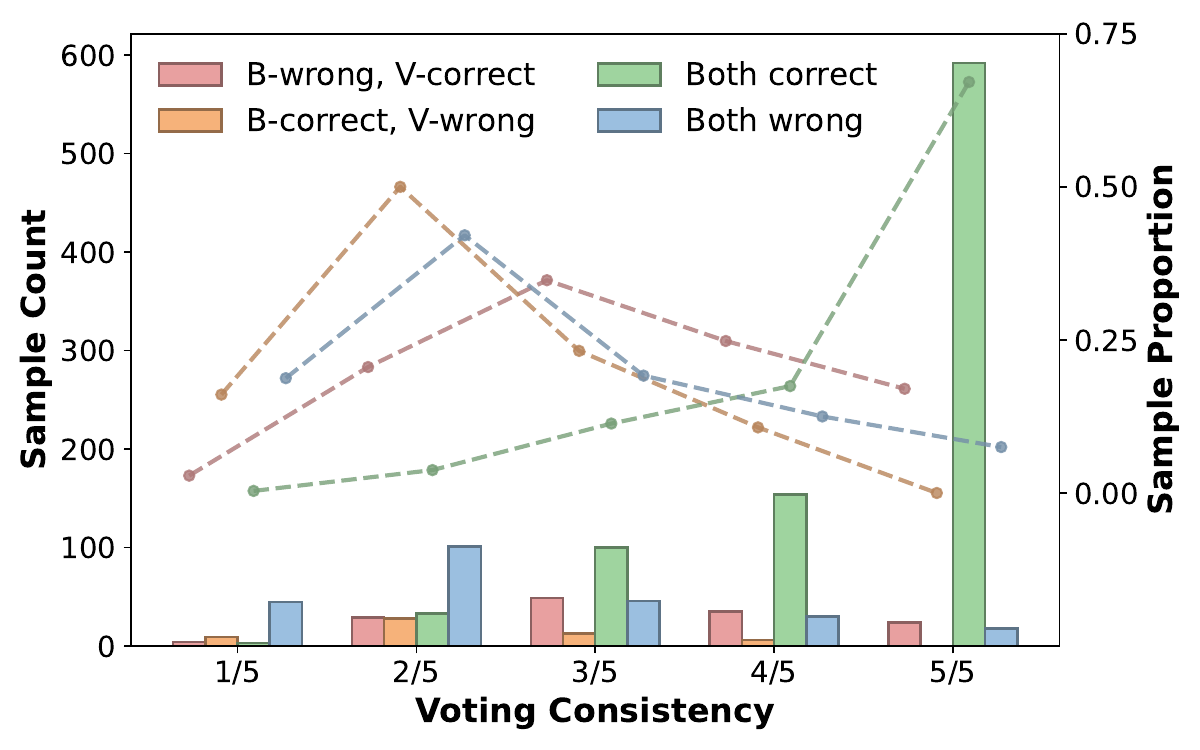} &
  \includegraphics[width=0.48\linewidth]{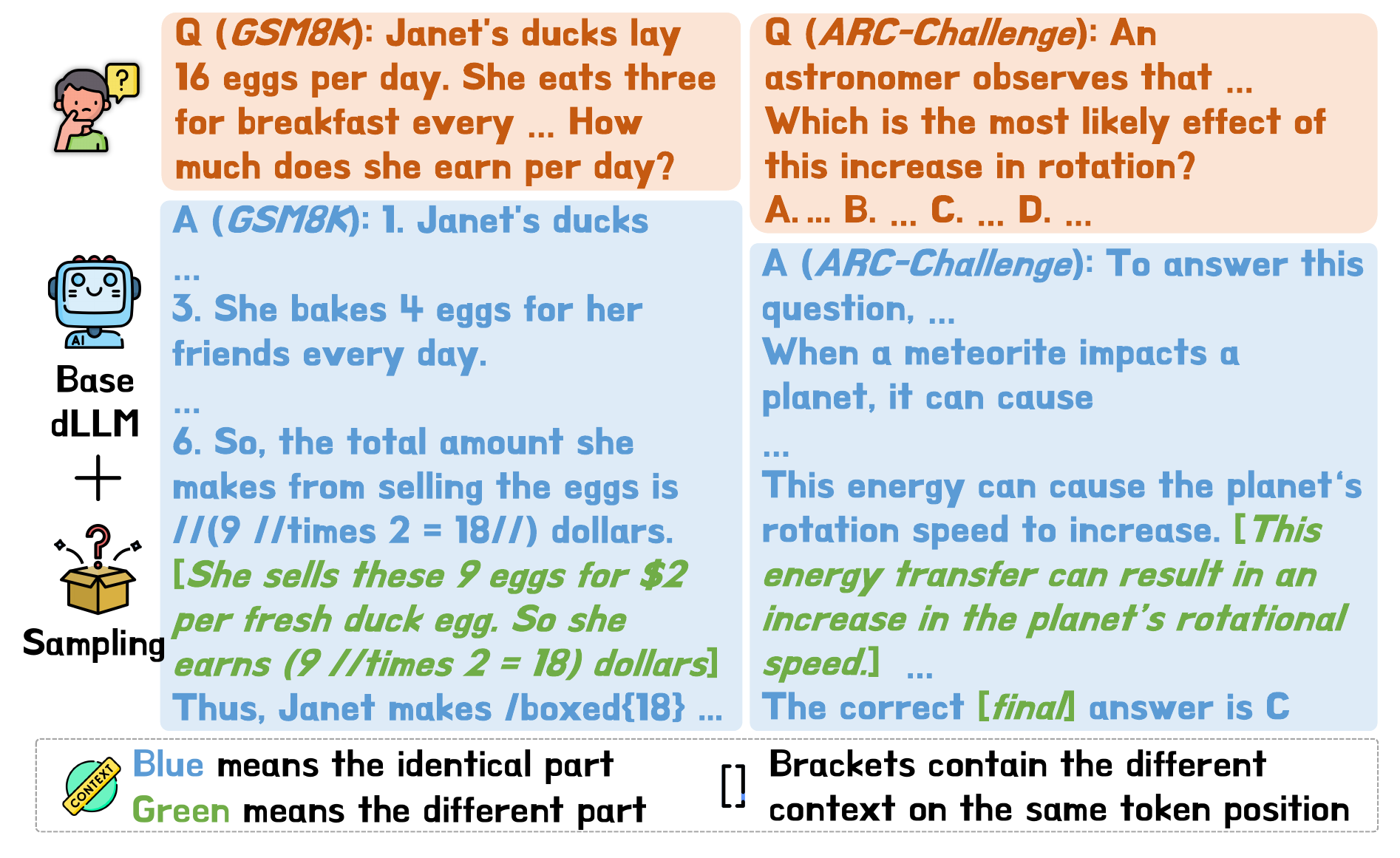} \\
  {\small (a) Distribution of voting consistency level} & {\small (b) Cases on token consistency} \\
\end{tabular}}
\vspace{-2mm}
\caption{(a) We report the distribution of voting consistency levels across different sample categories, defined by the correctness of the baseline and voting predictions. (b) We present two cases illustrating token-level redundancy in dLLM sampling ($5$ samples), drawn from GSM8K and ARC-C, respectively (zoom-in for more details).}
\label{fig:observations}
\vspace{-4mm}
\end{figure*}

\noindent \textbf{Properties of dLLMs Fit Parallel Test-Time Scaling.}
During inference, dLLMs update all masked positions in parallel, producing token predictions for the entire sequence at each iteration while selectively remasking uncertain positions. 
In contrast to prior work that focuses on single-run decoding~\citep{chen2025dparallel,wei2025accelerating}, our work primarily targets parallel test-time scaling, where multiple outputs are generated in parallel and subsequently aggregated into a final prediction. 
Across multiple sampling runs, the remasking mechanism enables the model to selectively reuse reliable context from previous iterations, refining earlier decisions. This property makes dLLMs a natural and scalable backbone for parallel test-time scaling. 
Additionally, the inherently parallel decoding process of dLLMs further reduces inference latency.


\noindent \textbf{Parallel Test-Time Scaling Can Work as an Alternative to Reinforcement Learning in dLLMs.}
RL can be viewed as selecting higher-quality trajectories from multiple samples and increasing their sampling probability, thereby improving sampling efficiency rather than fundamentally enhancing the model’s capacity. This perspective is also supported by recent work~\citep{yue2025does} in LLMs. As a major strategy of parallel test-time scaling, voting performs multiple sampling runs and selects the most consistent one as the final prediction. From this viewpoint, \textit{\textbf{voting can be regarded as a training-free alternative that approximates the effect of RL}}. As shown in Figure~\ref{fig:empirical-study}, we conduct experiments and further validate this claim in dLLMs: scaling test-time computation achieves performance comparable to advanced RL-based methods across both reasoning and general tasks. However, these significant performance gains come at the cost of increased inference overhead, which motivates us to improve efficiency.


\section{Methods}
\label{sec:methods}

In this section, we introduce \method to demonstrate the potential of simple voting strategies as a cost-effective alternative to RL in dLLMs. We first present the critical observation and then introduce the key designs of our \method. 





\subsection{Key Observations}
\label{sec:observation}

We conduct an empirical study using majority voting with entropy-threshold parallel decoding on LLaDA with GSM8K. Based on the results, we present the key observation that motivates us to leverage the remasking capability of dLLMs to further improve efficiency.

\begin{table}[t]
\centering
\caption{Results of NUPR@$k$ on various benchmarks under two generation lengths. To be specific, we record the tokens of $5$ samples and calculate the metric for each question.}
\label{tab:observation}
\vspace{-2mm}
\resizebox{0.995\linewidth}{!}{
\setlength{\tabcolsep}{1mm}
\begin{tabular}{lcccccc}
\toprule
\multirow{2}{*}{\textbf{Metric / Seq. Len.}} 
& \multicolumn{2}{c}{\textbf{GSM8K}} 
& \multicolumn{2}{c}{\textbf{MATH500}} 
& \multicolumn{2}{c}{\textbf{ARC-C}} \\
\cmidrule(l{4pt}r{4pt}){2-3} \cmidrule(l{4pt}r{4pt}){4-5} \cmidrule(l{4pt}r{4pt}){6-7}
& 128 & 256 
& 128 & 256 
& 128 & 256 \\
\midrule
NUPR@$2$
& 0.6077 & 0.5473 
& 0.4894 & 0.4383 
& 0.4612 & 0.4812 \\
NUPR@$3$
& 0.2109 & 0.1953 
& 0.1327 & 0.1142 
& 0.1562 & 0.2134 \\
\bottomrule
\vspace{-12mm}
\end{tabular}}
\end{table}

\noindent \textbf{Observation A.}
We partition samples into four categories based on whether the baseline and voting predictions are correct. We propose a metric called \textit{voting consistency level}, defined as the fraction of votes received by the most frequent answer. As shown in Figure~\ref{fig:observations}(a), we observe that samples correctly solved by both the baseline and voting methods are highly concentrated at high consistency levels, with $84.58\%$ falling into the $4$/$5$ or $5$/$5$ bins. In contrast, the remaining categories are predominantly associated with low consistency levels, with $66.07\%$ of samples receiving only $1$/$5$ or $2$/$5$ votes.
In summary, easier questions tend to exhibit higher voting consistency, indicating that sampling redundancy predominantly arises in such cases and can be reduced with minimal impact on performance.



\begin{table*}[ht]
\centering
\caption{Results on mathematical reasoning benchmarks with LLaDA-8B-Instruct. $*$ represents the results of reimplementation. $\dagger$ indicates that RFG requires an extra instruction-tuned or RL-enhanced model as the policy model. We report Pass@1 accuracy along with the corresponding step count. The involved RL-enhanced methods require in-domain training data.} 
\label{tab:llada-main-1}
\vspace{-2mm}
\resizebox{0.995\linewidth}{!}{
\setlength{\tabcolsep}{1mm}
\begin{tabular}{llcccccc}
\toprule
\multirow{2}{*}{\textbf{Method / Gen. Len.}} 
& \multirow{2}{*}{\textbf{Training Necessity}} 
& \multicolumn{3}{c}{\textbf{GSM8K (0-shot)}} 
& \multicolumn{3}{c}{\textbf{MATH500 (0-shot)}}\\ 
\cmidrule(l{4pt}r{4pt}){3-5} \cmidrule(l{4pt}r{4pt}){6-8}
&
&  128 & 256 & 512 
& 128 & 256 & 512 \\ 
\midrule
LLaDA-8B-Instruct & Pre-training & 
70.58\% / 128.0 & 76.72\% / 256.0  & 81.50\% / 512.0 & 
31.40\% / 128.0 & 35.00\% / 256.0  & 38.20\% / 512.0 \\ 
\midrule
\multicolumn{8}{c}{\textit{RL-Enhanced Models (in-domain)}} \\
\midrule 
d1 & SFT + RL & 
73.20\% / 64.0 & 81.10\% / 128.0 & 82.10\% / 256.0 & 
33.80\% / 64.0 & 38.60\% / 128.0 & 40.20\% / 256.0 \\  
wd1 & RL & 
- & 80.80\% / 128.0 & 82.30\% / 256.0 & 
- & 34.40\% / 128.0 & 39.00\% / 256.0 \\ 
IGPO & RL &
- & 83.60\% / 256.0 & - & 
- & 42.80\% / 256.0 & - \\ 
\midrule 
\multicolumn{8}{c}{\textit{Test-Time Scaling Strategies}} \\ 
\midrule 
Majority Voting & Training-free & 
76.72\% / 320.0 & 82.33\% / 640.0 & 86.28\% / 1280.0 & 
33.00\% / 320.0 & 36.80\% / 640.0 & 44.00\% / 1280.0 \\ 
HEX$^*$ & Training-free & 
80.14\% / 1600.0 & 85.75\% / 3200.0 & 88.78\% / 6400.0 & 
39.60\% / 1600.0 & 43.60\% / 3200.0 & 47.40\% / 6400.0 \\ 
RFG & FT \textit{or} RL$^\dagger$ &  
- & 81.30\% / 512.0 & - & 
- & 39.60\% / 512.0 & - \\
\rowcolor{gray!30} \method (ours) & Training-free & 
78.24\% / 170.4 & 83.78\% / 237.1 & 87.72\% / 289.6 & 
34.80\% / 292.1 & 40.20\% / 473.7 & 45.40\% / 701.2 \\ 
\bottomrule
\vspace{-4mm}
\end{tabular}}
\end{table*}

\begin{table*}[ht]
\centering
\caption{Results on scientific and general reasoning benchmarks with LLaDA-8B-Instruct. $*$ represents the results of reimplementation. We report Pass@1 accuracy along with the corresponding step count.} 
\label{tab:llada-main-2}
\vspace{-2mm}
\resizebox{0.995\linewidth}{!}{
\setlength{\tabcolsep}{1mm}
\begin{tabular}{llcccccc}
\toprule
\multirow{2}{*}{\textbf{Method / Gen. Len.}} 
& \multirow{2}{*}{\textbf{Training Necessity}} 
& \multicolumn{2}{c}{\textbf{ARC-C (0-shot)}} 
& \multicolumn{2}{c}{\textbf{GPQA (0-shot)}}
& \multicolumn{2}{c}{\textbf{MMLU (0-shot)}} \\ 
\cmidrule(l{4pt}r{4pt}){3-4} \cmidrule(l{4pt}r{4pt}){5-6} \cmidrule(l{4pt}r{4pt}){7-8}
& 
& 128 & 256 
& 128 & 256 
& 128 & 256 \\ 
\midrule
LLaDA-8B-Instruct & Pre-training & 
66.13\% / 128.0 & 74.66\% / 256.0  & 
25.00\% / 128.0 & 23.66\% / 256.0 & 
57.13\% / 128.0 & 58.04\% / 256.0 \\ 
\midrule 
\multicolumn{8}{c}{\textit{Test-Time Scaling Strategies}} \\ 
\midrule 
Majority Voting & Training-free & 
77.65\% / 320.0 & 81.31\% / 640.0 & 
22.54\% / 320.0 & 25.45\% / 640.0 & 
62.27\% / 320.0 & 63.11\% / 640.0 \\ 
HEX$^*$ & Training-free & 
82.67\% / 1600.0 & 83.87\% / 3200.0 & 
25.67\% / 1600.0 & 28.34\% / 3200.0 & 
64.28\% / 1600.0 & 64.75\% / 3200.0 \\ 
\rowcolor{gray!30} \method (ours) & Training-free & 
80.97\% / 265.4 & 83.36\% / 415.6 & 
28.57\% / 219.3 & 28.39\% / 555.7 & 
62.87\% / 222.1 & 62.87\% / 391.4 \\ 
\bottomrule
\vspace{-4mm}
\end{tabular}}
\end{table*}

\noindent \textbf{Observation B.} 
We further observe that for a given question, many token positions remain identical across multiple samples, as illustrated in Figure~\ref{fig:observations}(b). This token-level redundancy indicates substantial overlap among sampled sequences and reveals additional opportunities to reuse confident predictions while focusing resampling on a small subset of uncertain token positions.

We propose a new metric termed \textit{Non-Unique Position Rate at $k$ (NUPR@$k$)} to quantify how frequently this phenomenon occurs. Given $K$ sampled answers of equal length $N$ for the same question, NUPR@$k$ measures the fraction of token positions at which at least $k$ out of $K$ samples share an identical token. Formally, a token position is considered non-unique if at least $k$ out of the $K$ samples share the same token at that position, and NUPR@$k$ is computed as the average fraction of such positions across all tokens and questions. As shown in Table~\ref{tab:observation}, we report results on GSM8K, MATH500, and ARC-C under different generation lengths. We observe that NUPR@$2$ is consistently around 50\%, while NUPR@$3$ remains around 20\%, indicating substantial token-level redundancy.

\subsection{Remask Sampling}
\label{key-details-method}

Based on these observations (A \& B), we naturally connect the empirical findings with the inherent remasking capability of dLLMs and propose a simple yet effective remask sampling strategy (as shown in Figure~\ref{fig:overview}). We present the detailed implementation pipeline in Appendix~\ref{apx:detailed-algorithm}.

\noindent \textbf{Remask Sampling Strategy.}
Inspired by observation A, we stop sampling on time based on answer consistency (\textit{e.g.}, the first several generations yield the same answer). Furthermore, inspired by observation B, we propose our core design to further improve efficiency. Instead of generating a full response at each sampling action, we selectively remask tokens and continue sampling based on the consistency analysis. In particular, we retain tokens that exhibit agreement across samples, as well as tokens from samples with highly consistent predicted answers, and perform subsequent sampling conditioned on the preserved tokens. 

Additionally, in dLLMs, parallel decoding has been extensively explored and is a natural advantage of this paradigm. Following prior work~\citep{ben2025accelerated,chen2025dparallel,wei2025accelerating}, we adopt a simple entropy-threshold parallel decoding scheme that commits all token positions whose entropy falls below a predefined threshold~$\alpha$ at each denoising step. In our main experiments, we set $\alpha = 0.3$ and provide an ablation study on this threshold.

Finally, we stop sampling once the voting-based answer converges and adopt it as the final prediction.

\begin{table*}[ht]
\centering
\caption{Results on Dream-7B-Instruct. $*$ represents the results of reimplementation. $\dagger$ indicates that RFG requires an extra instruction-tuned or RL-enhanced model as the policy model. We set top-p to $0.6$ for the majority voting baseline. We report Pass@1 accuracy along with the corresponding step count.} 
\label{tab:dream-main}
\vspace{-2mm}
\resizebox{0.995\linewidth}{!}{
\setlength{\tabcolsep}{1mm}
\begin{tabular}{llcccccc}
\toprule
\multirow{2}{*}{\textbf{Method / Gen. Len.}} 
& \multirow{2}{*}{\textbf{Training Necessity}} 
& \multicolumn{2}{c}{\textbf{GSM8K (0-shot)}} 
& \multicolumn{2}{c}{\textbf{MATH500 (0-shot)}} 
& \multicolumn{1}{c}{\textbf{ARC-C (0-shot)}} 
& \multicolumn{1}{c}{\textbf{MMLU (0-shot)}} \\ 
\cmidrule(l{4pt}r{4pt}){3-4} \cmidrule(l{4pt}r{4pt}){5-6} \cmidrule(l{4pt}r{4pt}){7-7} \cmidrule(l{4pt}r{4pt}){8-8}
& 
&  128 & 256 
& 128 & 256 
& 128 
& 128 \\ 
\midrule
Dream-Base-7B$^*$ & - & 
68.92\% / 128.0 & 81.80\% / 256.0  & 
33.00\% / 128.0 & 42.20\% / 256.0  & 
66.63\% / 128.0 & 
64.83\% / 128.0 \\ 
\midrule 
\multicolumn{8}{c}{\textit{Test-Time Scaling Strategies}} \\ 
\midrule 
Majority Voting & Training-free & 
72.33\% / 320.0 & 84.53\% / 640.0 & 
34.20\% / 320.0 & 44.00\% / 640.0 & 
77.73\% / 320.0 & 
68.51\% / 320.0 \\ 
HEX$^*$ & Training-free &
77.48\% / 1600.0 & 87.04\% / 3200.0 & 
43.60\% / 1600.0 & 50.60\% / 3200.0 & 
88.14\% / 1600.0 & 
69.87\% / 1600.0 \\ 
RFG & FT \textit{or} RL$^\dagger$ &  
- & 82.10\% / 512.0 & 
- & 46.40\% / 512.0 & 
- & 
- \\
\rowcolor{gray!30} \method (ours) & Training-free & 
75.44\% / 206.2 & 86.96\% / 290.4 & 
42.20\% / 320.7 & 48.80\% / 498.5 & 
86.09\% / 221.9 &
69.80\% / 119.6 \\ 
\bottomrule
\vspace{-4mm}
\end{tabular}}
\end{table*}

\section{Experiments}
\label{sec:experiments}

\begin{figure*}[t]
\centering
\includegraphics[width=\linewidth]{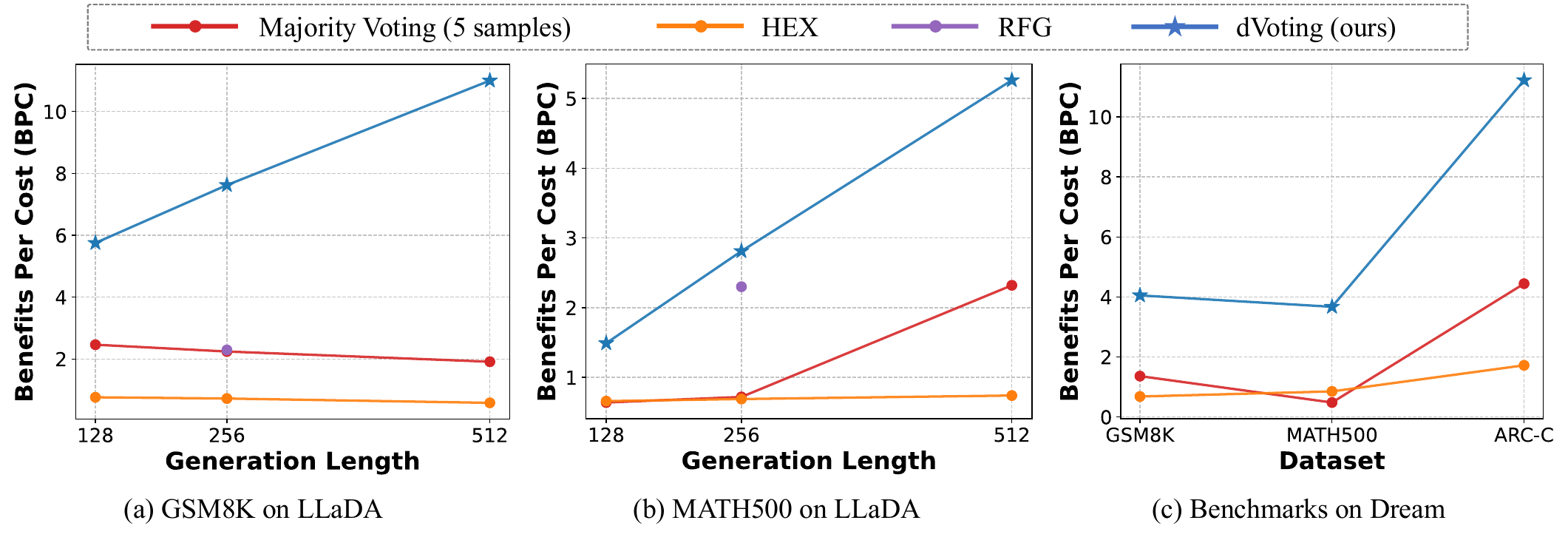} \\
\vspace{-2mm}
\caption{Comparison of performance-efficiency trade-off between \method and other test-time scaling baselines. We present the results on the LLaDA model for GSM8K and MATH500 in (a) and (b), respectively. We present the results on the Dream model for GSM8K, MATH500, and ARC-C under $128$ generation length in (c). We mark our method with a star to distinguish it from other methods.}
\label{fig:pareto-frontier}
\vspace{-4mm}
\end{figure*}

\subsection{Experimental Setups}
\label{sec:experimental-setup}

\noindent \textbf{Baselines.}
We compare against four sampling baselines: the original results, majority voting, and two related methods, HEX~\citep{lee2025test} and RFG~\citep{chen2025rfg}. Specifically, for the original performance, we adopt the semi-autoregressive decoding strategy following the original paper; for majority voting ($5$ samples), we follow the configuration used in HEX. For HEX and RFG, we report either our reimplementations or the results from the original papers; the two methods require $25$ and $2$ samples, respectively. Additionally, we include the original results reported in the paper for three RL methods: d1~\citep{zhao2025d1}, wd1~\citep{tang2025wd1}, and IGPO~\citep{zhao2025inpainting}.

\noindent \textbf{Inference Details.}
We follow the official inference settings to evaluate the original models from both the LLaDA series~\citep{nie2025large,zhu2025llada} and Dream~\citep{ye2025dream}. For the baselines, we report the results from the original paper or our reimplementation, with the latter following the configurations specified in the original paper. 
We apply full denoising steps ($N$) for evaluating the original models and half denoising steps ($N/2$) for baselines (\textit{e.g.}, majority voting and HEX) to generate a sequence of length $N$.
We set the temperature to $0.6$ for our proposed \method and conduct experiments under standard generation lengths ($128$, $256$, $512$), with corresponding block sizes of $8$, $16$, and $32$, respectively, following the semi-autoregressive strategy for models from the LLaDA series. For the Dream model, we additionally implement a semi-autoregressive decoding strategy, as it is not supported by the original generation pipeline, while all other settings follow those used for the LLaDA series. We set an upper bound $n = 5$ on the total number of samples for the proposed \method.

\noindent \textbf{Evaluation Details.}
We conduct extensive experiments across various benchmarks covering a broad range of reasoning tasks, including mathematical reasoning, scientific reasoning, and general high-level reasoning, using GSM8K~\citep{cobbe2021gsm8k}, MATH500~\citep{lightman2023let}, ARC-C~\citep{clark2018think}, MMLU~\citep{hendryckstest2021}, and GPQA~\citep{rein2024gpqa}. Additionally, for models from the LLaDA (LLaDA-8B-Instruct and LLaDA-1.5) and Dream (Dream-7B-Instruct), we follow the simple-eval framework\footnote{\url{https://github.com/openai/simple-evals}} for zero-shot evaluation and prompt the models to generate step-by-step reasoning. We implement the voting procedure within the same framework to process all candidate answers.

\subsection{Main Results}
\label{sec:main-results}

\noindent \textbf{Results on the Native dLLM (LLaDA).}   
We report the results of LLaDA-8B-Instruct in Tables \ref{tab:llada-main-1} and \ref{tab:llada-main-2}. Compared to the original model and most in-domain RL methods, our \method achieves consistent and substantial performance gains across different benchmarks and generation lengths: $6.22\%\text{–}7.66\%$ on GSM8K, $4.40\%\text{–}7.20\%$ on MATH500, $3.16\%\text{-}14.84\%$ on ARC-C, $3.57\%\text{-}4.73\%$ on GPQA, and $4.83\%\text{-}5.74\%$ on MMLU. Against test-time scaling baselines, \method attains strong performance with the fewest steps, delivering $1.1\text{–}4.4\times$ speedup over majority voting, $1.1\text{–}2.2\times$ over RFG, and $5.5\text{–}22.1\times$ over HEX.

\begin{figure*}[ht]
\centering
  \includegraphics[width=\linewidth]{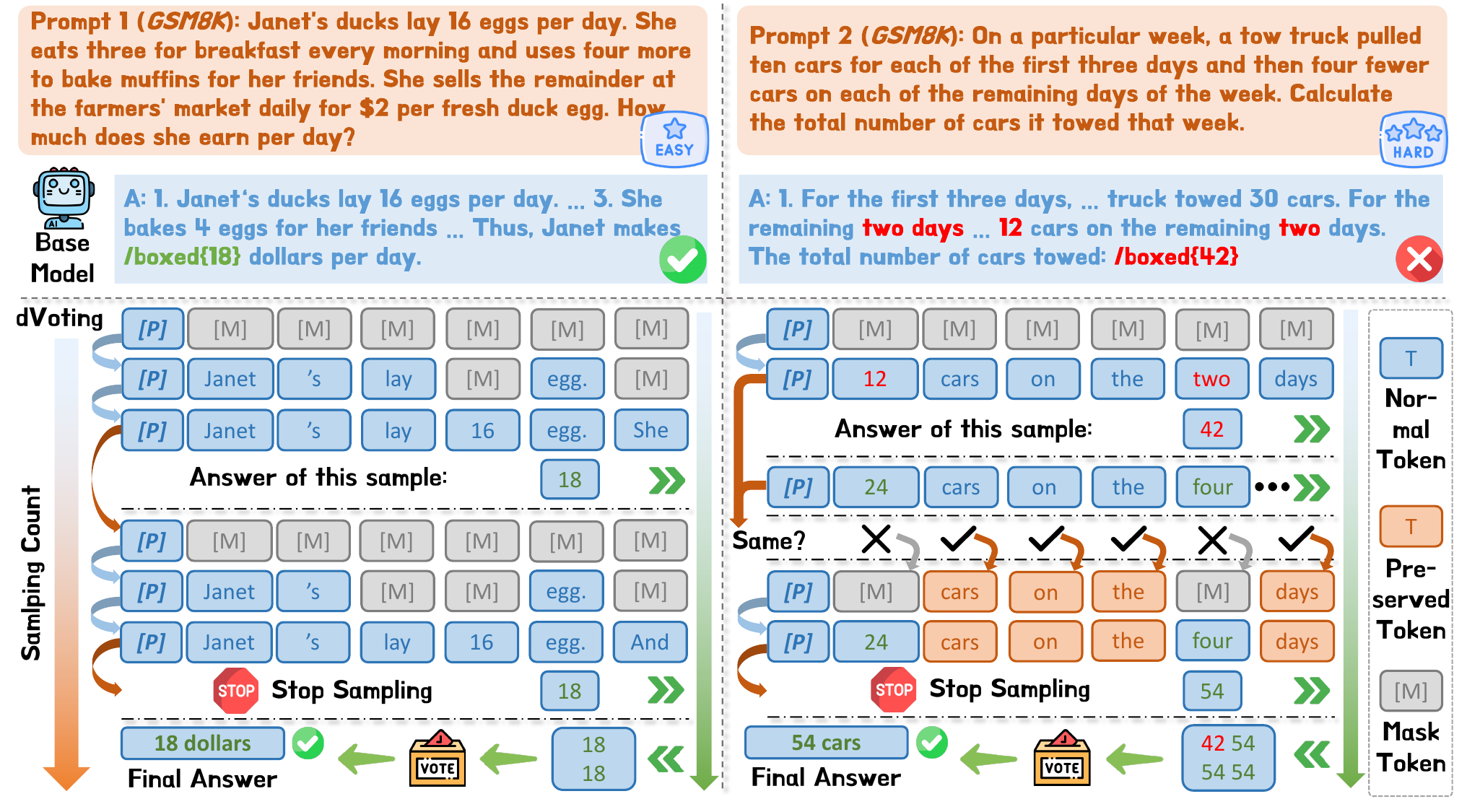} \\
\vspace{-2mm}
\caption{Case visualization. We present two cases where the original model yields a correct and an incorrect answer, respectively. Prompt-2 fails by incorrectly assuming a five-day week, while our method iteratively refines this part to produce the correct answer.}
\label{fig:case-analysis}
\vspace{-2mm}
\end{figure*}

\begin{table*}[ht]
\centering
\caption{Results on RL-enhanced model (LLaDA-1.5). $*$ represents the results of reimplementation. $\dagger$ indicates that RFG requires an extra instruction-tuned or RL-enhanced model as the policy model. We report Pass@1 accuracy along with the corresponding step count.} 
\label{tab:rl-enhanced-results}
\vspace{-2mm}
\resizebox{0.995\linewidth}{!}{
\setlength{\tabcolsep}{1mm}
\begin{tabular}{llcccccccc}
\toprule
\multirow{2}{*}{\textbf{Method / Gen. Len.}} 
& \multirow{2}{*}{\textbf{Training Necessity}} 
& \multicolumn{2}{c}{\textbf{GSM8K (0-shot)}} 
& \multicolumn{2}{c}{\textbf{MATH500 (0-shot)}}
& \multicolumn{1}{c}{\textbf{ARC-C (0-shot)}}
& \multicolumn{1}{c}{\textbf{MMLU (0-shot)}}\\ 
\cmidrule(l{4pt}r{4pt}){3-4} \cmidrule(l{4pt}r{4pt}){5-6} \cmidrule(l{4pt}r{4pt}){7-7} \cmidrule(l{4pt}r{4pt}){8-8}
&
& 128 & 256 
& 128 & 256 
& 128 
& 128 \\ 
\midrule
LLaDA-1.5$^*$ & - & 
72.48\% / 128.0 & 78.01\% / 256.0  & 
30.60\% / 128.0 & 33.40\% / 256.0  & 
68.52\% / 128.0 & 
58.27\% / 128.0 \\ 
\midrule 
\multicolumn{8}{c}{\textit{Test-Time Scaling Strategies}} \\ 
\midrule 
Majority Voting & Training-free & 
78.47\% / 320.0 & 84.69\% / 640.0 & 
32.00\% / 320.0 & 41.00\% / 640.0 & 
79.69\% / 320.0 & 
62.91\% / 320.0 \\ 
HEX~\citep{lee2025test}$^*$ & Training-free & 
82.41\% / 1600.0 & 86.50\% / 3200.0 & 
39.60\% / 1600.0 & 43.40\% / 3200.0 & 
86.69\% / 1600.0 & 
64.47\% / 1600.0 \\ 
RFG~\citep{chen2025rfg} & FT \textit{or} RL$^\dagger$ &  
- & 82.10\% / 512.0 & 
- & - & 
- & 
- \\ 
\rowcolor{gray!30} \method (ours) & Training-free & 
79.15\% / 158.8 & 84.53\% / 223.1 & 
34.40\% / 280.1 & 41.20\% / 443.1 & 
82.63\% / 267.7 & 
63.10\% / 224.0 \\ 
\bottomrule 
\vspace{-4mm} 
\end{tabular}} 
\end{table*}

\noindent \textbf{Results on the AR-initialized dLLM (Dream).}
As shown in Table \ref{tab:dream-main}, our \method also consistently achieves significant performance improvements on the Dream model across all benchmarks while using the fewest steps among all test-time scaling baselines. Specifically, \method delivers a $1.0\text{–}2.7\times$ speedup over majority voting, a $1.0\text{–}1.8\times$ speedup over RFG, and a $5.0\text{–}13.4\times$ speedup over HEX.

\begin{table*}[ht]
\centering
\caption{Ablation on sampling upper bound $n$. We evaluate sampling upper bounds of $1$, $5$, $9$, $13$, and $17$ while keeping all other hyperparameters fixed, and report Pass@1 accuracy and corresponding step count. We provide the majority voting baseline as a reference.}
\label{tab:ablation-upper-bound}
\vspace{-2mm}
\resizebox{0.995\linewidth}{!}{
\setlength{\tabcolsep}{1mm}
\begin{tabular}{lccccccc}
\toprule
\multirow{2}{*}{\textbf{Dataset / Gen. Len.}}
& \multicolumn{2}{c}{\textbf{Baselines}} 
& \multicolumn{5}{c}{\textbf{\method}} \\ 
\cmidrule(l{4pt}r{4pt}){2-3} \cmidrule(l{4pt}r{4pt}){4-8}
& \textbf{LLaDA-8B-Instruct} 
& \textbf{Majority Voting} 
& $n$=1 
& $n$=5 
& $n$=9
& $n$=13
& $n$=17 \\
\midrule
GSM8K / 128 & 
70.58\% / 128.0 &
76.72\% / 320.0 &
70.13\% / 58.2  &  
78.24\% / 170.4 &
79.68\% / 348.6 &  
80.89\% / 515.5 &  
80.89\% / 683.2 \\ 
GSM8K / 256 & 
76.72\% / 256.0 &
82.33\% / 640.0 &
77.55\% / 85.3  &
83.78\% / 237.1 &  
85.97\% / 477.3 &  
86.13\% / 708.6 &  
86.20\% / 937.7 \\ 
\midrule
MATH500 / 128 & 
31.40\% / 128.0 &
33.00\% / 320.0 &
27.60\% / 72.0  &
34.80\% / 292.1 &  
36.40\% / 573.3 &  
37.60\% / 837.9 &  
39.20\% / 1105.0 \\ 
MATH500 / 256 & 
35.00\% / 256.0 &
36.80\% / 640.0 &
36.20\% / 120.5 &
40.20\% / 473.7 &  
42.20\% / 928.2 &  
43.20\% / 1350.7 &  
43.20\% / 1779.3 \\ 
\bottomrule
\vspace{-4mm}
\end{tabular}}
\end{table*}

\begin{table*}[ht]
\centering
\caption{Ablation on block size. We evaluate block sizes of $4$, $8$, $16$, $32$, and $64$ while keeping all other hyperparameters fixed, and report Pass@1 accuracy along with the corresponding step count. We provide the majority voting baseline as a reference.}
\label{tab:ablation-block-size}
\vspace{-2mm}
\resizebox{0.995\linewidth}{!}{
\setlength{\tabcolsep}{1mm}
\begin{tabular}{lccccccc}
\toprule
\multirow{2}{*}{\textbf{Dataset / Gen. Len.}}
& \multicolumn{2}{c}{\textbf{Baselines}} 
& \multicolumn{5}{c}{\textbf{\method}} \\ 
\cmidrule(l{4pt}r{4pt}){2-3} \cmidrule(l{4pt}r{4pt}){4-8}
& \textbf{LLaDA-8B-Instruct} 
& \textbf{Majority Voting} 
& block size=4 
& block size=8 
& block size=16
& block size=32
& block size=64 \\
\midrule
GSM8K / 128 & 
70.58\% / 128.0 &
76.72\% / 320.0 &
79.23\% / 168.6 &  
78.24\% / 170.4 &  
77.71\% / 176.1 &  
78.01\% / 177.8 &  
75.82\% / 184.3 \\ 
GSM8K / 256 & 
76.72\% / 256.0 &
82.33\% / 640.0 &
83.70\% / 227.2 & 
84.23\% / 231.5 & 
83.78\% / 237.1 &
84.23\% / 238.8 &
84.76\% / 238.6 \\
\midrule
MATH500 / 128 & 
31.40\% / 128.0 &
33.00\% / 320.0 &
35.00\% / 281.0 & 
34.80\% / 292.1 &  %
34.00\% / 296.4 &  %
34.00\% / 297.9 &  %
34.80\% / 308.8 \\ %
MATH500 / 256 & 
35.00\% / 256.0 &
36.80\% / 640.0 &
40.00\% / 448.1 &  %
40.00\% / 469.8 &  %
40.20\% / 473.7 &  %
40.40\% / 478.5 &  %
40.40\% / 463.7 \\ %
\bottomrule
\vspace{-4mm}
\end{tabular}}
\end{table*}
\begin{table*}[!h]
\centering
\caption{Ablation on entropy threshold $\alpha$. We evaluate entropy thresholds of $0.1$, $0.3$, $0.5$, $0.7$, and $0.9$ while keeping all other hyperparameters fixed, and report Pass@1 accuracy and corresponding step count. We provide the majority voting baseline as a reference.}
\label{tab:ablation-entropy-threshold}
\vspace{-2mm}
\resizebox{0.995\linewidth}{!}{
\setlength{\tabcolsep}{1mm}
\begin{tabular}{lccccccc}
\toprule
\multirow{2}{*}{\textbf{Dataset / Gen. Len.}}
& \multicolumn{2}{c}{\textbf{Baselines}} 
& \multicolumn{5}{c}{\textbf{\method}} \\ 
\cmidrule(l{4pt}r{4pt}){2-3} \cmidrule(l{4pt}r{4pt}){4-8}
& \textbf{LLaDA-8B-Instruct} 
& \textbf{Majority Voting} 
& $\alpha$=0.1 
& $\alpha$=0.3 
& $\alpha$=0.5
& $\alpha$=0.7
& $\alpha$=0.9 \\
\midrule
GSM8K / 128 & 
70.58\% / 128.0 &
76.72\% / 320.0 &
78.24\% / 213.7 &  
78.24\% / 170.4 & 
77.48\% / 152.6 &  
77.18\% / 140.4 & 
74.91\% / 136.4 \\ 
GSM8K / 256 & 
76.72\% / 256.0 &
82.33\% / 640.0 &
84.76\% / 299.9 &  
83.78\% / 237.1 &  
83.70\% / 206.1 &  
82.48\% / 188.6 &  
80.14\% / 189.4 \\ 
\midrule
MATH500 / 128 & 
31.40\% / 128.0 &
33.00\% / 320.0 &
34.80\% / 345.7 &  
34.80\% / 292.1 & 
33.80\% / 260.9 &  
33.60\% / 242.5 &  
31.40\% / 221.4 \\ 
MATH500 / 256 & 
35.00\% / 256.0 &
36.80\% / 640.0 &
42.60\% / 565.4 &  
40.20\% / 473.7 &  
40.80\% / 411.0 &  
39.20\% / 378.0 & 
35.20\% / 361.5 \\ 
\bottomrule
\vspace{-4mm}
\end{tabular}}
\end{table*}

\noindent \textbf{\method Pushes Performance–Efficiency Pareto Frontier Forward.}
We propose a unified metric, denoted as \textit{benefits per cost (BPC)}, to quantify the performance–efficiency trade-off in the context of test-time scaling. Specifically, the metric measures the performance gain per unit of additional test-time cost, computed as the improvement over the original model normalized by the corresponding increase in the total number of denoising steps\footnote{We present an example (GSM8K on LLaDA-8B-Instruct under $128$ generation length): 
$\frac{78.24\% - 70.58\%}{170.4~/~128.0} = 5.75$.}. As shown in Figure~\ref{fig:pareto-frontier}, our \method achieves superior performance–efficiency trade-off across all settings. Based on the LLaDA results, our \method consistently achieves larger BPC improvements than other test-time scaling baselines as the generation length increases, demonstrating its clear advantage in long-generation length settings.

\noindent \textbf{Generalization of \method on RL-enhanced Models.}
We next evaluate the generalization of \method on RL-enhanced models, such as LLaDA-1.5 fine-tuned with Variance-Reduced Preference Optimization (VRPO)~\citep{zhu2025llada}. As shown in Table~\ref{tab:rl-enhanced-results}, we conduct experiments across four benchmarks. The results demonstrate that \method consistently brings significant performance improvements on RL-enhanced models while incurring only minimal additional inference cost, further confirming the generalization and effectiveness of our method.

\subsection{Visualization}
\label{sec:qualitative-results}

Figure~\ref{fig:case-analysis} illustrates two GSM8K cases where LLaDA produces a correct and an incorrect prediction, respectively. For Prompt~1, the base model already yields the correct answer, and our \method quickly identifies that no further remasking is needed, terminating sampling early. For Prompt~2, the base model fails, while our method performs five sampling iterations and aggregates the results via voting to recover the correct answer. Overall, our proposed \method can adapt its computation according to problem difficulty automatically, efficiently solving simple cases while allocating more inference effort to harder ones to improve both efficiency and performance. We provide more cases in Appendix~\ref{apx:case-study}.


\subsection{Ablation Study}
\label{sec:diagnostic-experiments}

We provide comprehensive ablation studies on sampling upper bound $n$, block size, and entropy threshold $\alpha$ on LLaDA, evaluated on GSM8K and MATH500.

\noindent \textbf{Ablation on Sampling Upper Bound.}
We first ablate the sampling upper bound $n$ under two generation lengths ($128$ and $256$). As shown in Table~\ref{tab:ablation-upper-bound}, performance consistently improves with larger $n$ across both benchmarks and both generation lengths, until reaching a saturation point. This demonstrates that \method can effectively scale test-time computation to boost performance, consistent with the test-time scaling law stated in prior work~\citep{snell2024scaling}.

\noindent \textbf{Ablation on Block Size.}
We next ablate the block size, with results summarized in Table~\ref{tab:ablation-block-size}. Across block sizes ranging from $4$ to $64$, \method consistently outperforms both the original results and the majority voting baseline, indicating that its performance is robust to this hyperparameter and further demonstrating the effectiveness of our approach.

\noindent \textbf{Ablation on Entropy Threshold.}
We further conduct an ablation study on the entropy threshold $\alpha$. As shown in Table~\ref{tab:ablation-entropy-threshold}, \method maintains strong performance and a favorable efficiency trade-off across a broad range of threshold values (\textit{e.g.}, $0.1\text{-}0.7$). A noticeable performance drop is observed only when the threshold is set excessively high, such as $\alpha$=$0.9$, where \method still outperforms the original results. Moreover, the results follow a test-time scaling trend, with increased computation leading to improved performance. Overall, this ablation study further demonstrates the robustness and effectiveness of our proposed method.

\section{Conclusion}
\label{sec:conclusion}

In this work, we address the challenge of the high inference cost incurred by test-time scaling in dLLMs. We first identify a key empirical insight: repeated tokens frequently emerge across multiple samples for the same prompt. Building on this observation, we leverage the remasking mechanism of dLLMs and propose \method, a fast voting strategy tailored to dLLMs. Extensive evaluations demonstrate that our method consistently enhances reasoning performance with modest extra inference cost, achieving a leading performance–efficiency trade-off. Overall, our work provides a foundation for future test-time scaling and unlocks the inference-time potential of dLLMs.




\section*{Impact Statement}

This work investigates efficient test-time scaling for dLLMs by proposing a simple voting strategy that leverages the remasking mechanism to focus computation on uncertain tokens. The approach improves reasoning performance while significantly reducing inference cost, making test-time scaling more practical under limited computational budgets. By improving inference efficiency, the proposed approach reduces computational overhead and resource consumption, thereby making large-scale model deployment more feasible in practice. The method does not introduce additional ethical or societal concerns beyond those commonly associated with large language models.

\bibliographystyle{icml2026}

\clearpage
\appendix
\onecolumn
\setcounter{figure}{0}
\setcounter{table}{0}
\renewcommand{\thefigure}{A\arabic{figure}}
\renewcommand{\thetable}{A\arabic{table}}
\section*{Appendix}
\label{apx:apx}

In Appendix~\ref{apx:detailed-algorithm}, we present the detailed process of our \method. Appendix~\ref{apx:case-study} includes additional case analysis to further demonstrate the effectiveness of our \method. We further discuss future works in Appendix~\ref{apx:future-work}. Finally, we present the LLM usage statement in Appendix~\ref{apx:llm-usage}.

\vspace{-0.2cm}
\startcontents[appendices]
\printcontents[appendices]{l}{1}{\setcounter{tocdepth}{3}}

\section{The Details of \method}
\label{apx:method-details}

\subsection{The Detailed Algorithm of the Consistency-Guided Designs}
\label{apx:detailed-algorithm}

We provide the detailed algorithm for the pipeline of our \method in Algorithm~\ref{alg:dtts-process}. The entire pipeline is simple and easy to implement. We implement this pipeline with the PyTorch library~\citep{paszke2019pytorch}.

\begin{algorithm*}[ht]
\caption{Detailed Process of dVoting}
\label{alg:dtts-process}
\begin{algorithmic}[1]
\REQUIRE Prompt $p$, dLLM $f_\theta$, maximum samples $n$, generation length $L$
\STATE $X_{\text{all}} \leftarrow \emptyset$ \COMMENT{store all sampled generations}

\FOR{$i = 1$ to $n$}

    \STATE Initialize sequence $x$ with prompt tokens and masked generation positions
    \STATE Initialize a Boolean array $m \in \{0,1\}^L$ representing whether remask or not

    \IF{$X_{\text{all}} \neq \emptyset$}
        \FOR{each generation position $t$}
            \STATE Compute the token consistency score at position $t$ from previous samples
            \IF{the score at position $t$ does not satisfy the remasking criterion}
                \STATE Fix the token at position $t$
                \STATE Set $m[t] \leftarrow \text{False}$
            \ELSE
                \STATE Mask the token at position $t$
                \STATE Set $m[t] \leftarrow \text{True}$
            \ENDIF
        \ENDFOR
    
        \STATE Evaluate global token-level consistency across all positions
        \IF{the consistency scores indicate that no token requires further remasking}
            \STATE Fix all tokens
            \STATE Set $m[t] \leftarrow \text{False}$ for all $t \in \{1,\dots,L\}$
        \ENDIF
    \ENDIF
    
    \IF{all entries of $m$ are \text{False}}
        \STATE \textbf{break} \COMMENT{\textit{Stop sampling here!}}
    \ENDIF
    
    \IF{there exist masked tokens in $x$}
        \STATE Perform entropy-threshold-based parallel decoding with a semi-autoregressive strategy \COMMENT{\textit{Details omitted as this is a typical strategy!}}
    \ENDIF
    
    \STATE Append the completed sequence $x$ to $X_{\text{all}}$
    
\ENDFOR

\STATE \textbf{return} $X_{\text{all}}$

\STATE Perform voting on candidate answers in $X_{\text{all}}$ to get final prediction \COMMENT{\textit{Voting part is here!}}
\end{algorithmic}
\end{algorithm*}

\section{Case Analysis}
\label{apx:case-study}

\begin{figure*}[ht]
\centering
  \includegraphics[width=\linewidth]{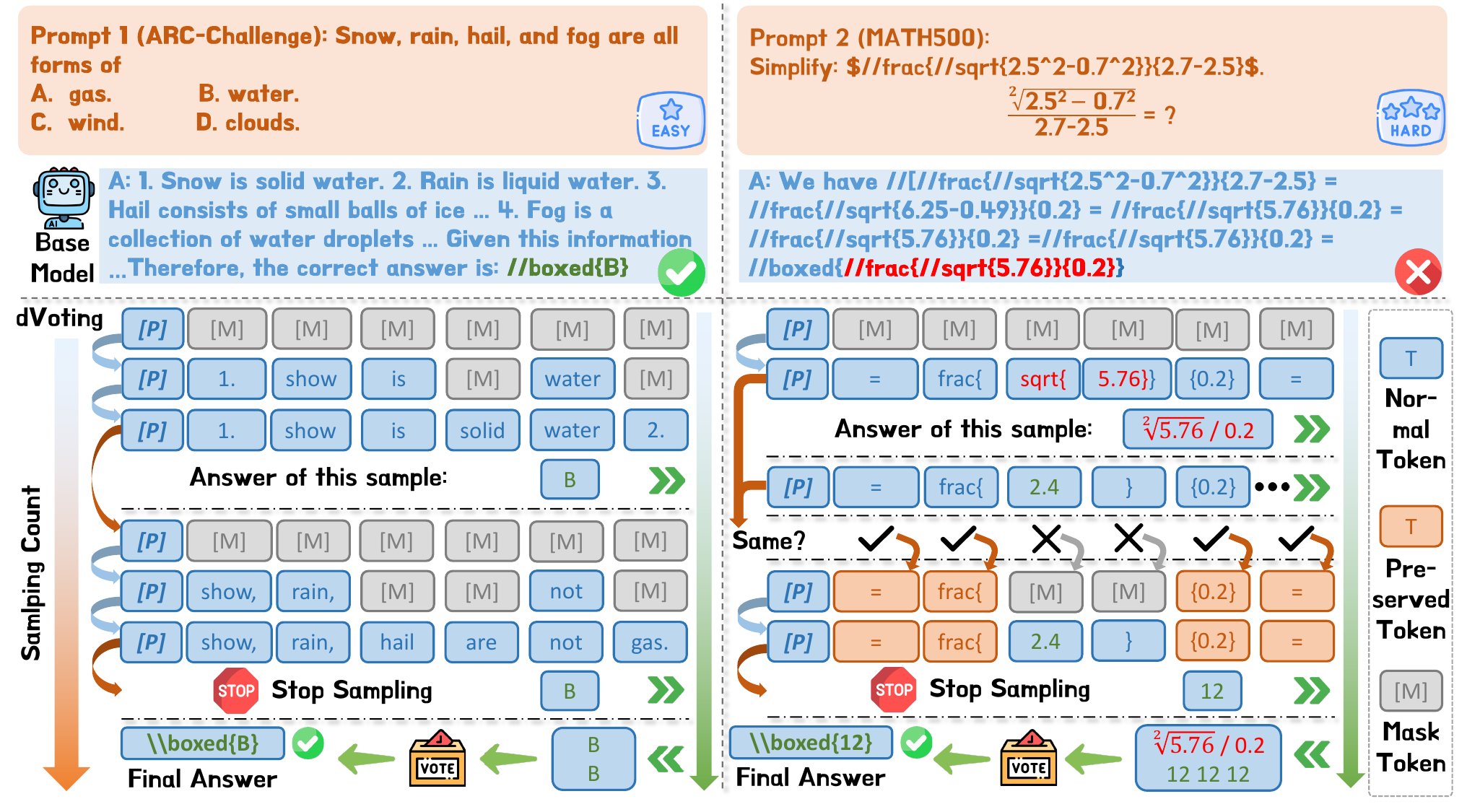} \\
\vspace{-2mm}
\caption{Additional case visualization.}
\label{fig:case-analysis-apx}
\vspace{-2mm}
\end{figure*}

As shown in Figure~\ref{fig:case-analysis-apx}, we present additional examples from ARC-C and MATH500, further demonstrating that our method quickly produces correct answers on simple problems and refines initially incorrect answers on more challenging ones. 
Prompt 1 is an easy ARC-C question, where \method early-stops after two samplings upon obtaining consistent answers. In contrast, Prompt 2 is a harder MATH500 problem. While the base model outputs an incorrect answer in the first sampling, \method refines key tokens through multiple samplings and stops after four sampling runs when a consistent answer is observed, yielding the correct result.

\section{Future Work}
\label{apx:future-work}

Our method and experiments primarily focus on the language modality. As an emerging architecture, diffusion language models have demonstrated performance comparable to autoregressive models in language tasks, while also showing promise for extension to other modalities and tasks~\citep{li2025every}, such as multimodal question answering~\citep{yu2025dimple,you2025llada,yang2025mmada,feng2025can,yu2025discrete}. Exploring parallel test-time scaling in these broader settings is a promising direction for future work. Moreover, parallel test-time scaling is simple to implement (easy to code) and computationally lightweight (easy to run), making it easy to adopt and follow. In practice, \method can be applied to resource-constrained and cost-sensitive deployment settings, such as on-device reasoning and large-scale inference services.

\section{Large Language Model Usage Statement}
\label{apx:llm-usage}

Large language models were used only for minor language editing, such as improving grammar, clarity, and formatting. They did not contribute to the development of ideas, methods, algorithms, code, experiments, figures, or analyses. All technical work and experimental results were produced by the authors. Any model-assisted text was manually reviewed, and the use of LLMs does not affect the reproducibility of the work.

\end{document}